\documentclass{article}
\usepackage[utf8]{inputenc}
\usepackage{url}
\usepackage{graphicx}
\usepackage{footnote}
\makesavenoteenv{tabular}
\usepackage{hyperref}
\title{Satellite Image Semantic Segmentation}
\author{Eric Guérin$^1$, Killian Oechslin$^1$, Christian Wolf$^1$, Benoît Martinez$^2$\\[1em]$^1$ CNRS LIRIS - INSA Lyon\\$^2$ Ubisoft}
\date{September 2021}

\begin{document}

\maketitle
\begin{centering}
\includegraphics[width=0.5\linewidth]{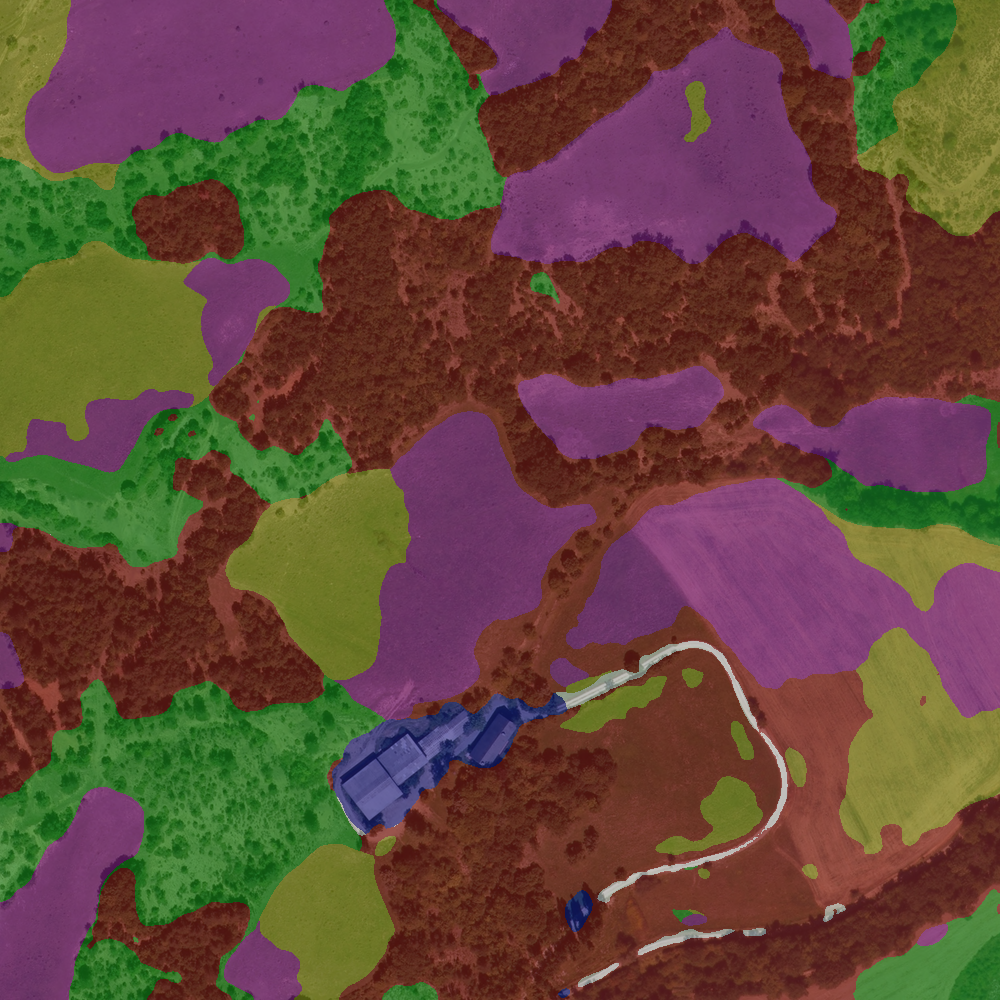}\\
\includegraphics[width=\linewidth]{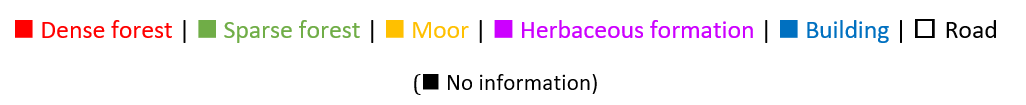}
\end{centering}

\paragraph*{Abstract}
In this paper, we propose a method for the automatic semantic segmentation of satellite images into six classes (sparse forest, dense forest, moor, herbaceous formation, building, and road). We rely on Swin Transformer architecture and build the dataset from IGN open data. We report quantitative and qualitative segmentation results on this dataset and discuss strengths and limitations. The dataset and the trained model are made publicly available.

\section{Introduction}
Virtual worlds in the context of digital entertainment need to be vast and realistic. These two factors force industries to resort to 
using artists massively. In the same time, more and more geographic data such as digital satellite photography become publicly available. 
Unfortunately, this data is rarely segmented and cannot be used directly. In the context 
of the ANR project Ampli \footnote{\url{https://projet.liris.cnrs.fr/ampli/}}, we aim at making the task of virtual worlds authoring easier by providing
a way to segment satellite images into six basic landcover classes. The segmentation method we use is Swin Transformer \cite{Liu2021} (section \ref{sec:method}) and we 
build the dataset from IGN public data (section \ref{sec:data}). The obtained results are very promising (section \ref{sec:results}) and the trained model is made publicly available together with the training dataset.

\section{Swin Transformer Semantic Segmentation}
\label{sec:method}
Swin Transformer \cite{Liu2021} is a general purpose computer vision backbone that has been proven very efficient and recently at the top of the state-of-the-art for image classification, object detection, and semantic segmentation. Its architecture based on Shifted WINdows makes it robust against
scale variability while keeping linear efficiency with respect to the number of pixels. The Shift Windows concept consists in having a window shifted by
half of its size in order to limit the self-attention computation to non-overlapping local windows while keeping possible to have cross-window
connections. 

In our experiments, we use an implementation \footnote{\url{https://github.com/SwinTransformer/Swin-Transformer-Semantic-Segmentation}} based on mmsegmentation~\cite{mmseg2020}. 

\section{Data Preparation and Setup}
\label{sec:data}
\subsection{Dataset sources}
To train and test the model, we used open data provided by IGN \footnote{\url{https://geoservices.ign.fr/telechargement}} which concerns French departments (Hautes-Alpes in our case). The following datasets have been used to extract the different layers:
\begin{itemize}
\item BD Ortho for the satellite images
\item BD Foret v2 for vegetation data
\item BD Topo for buildings and roads
\end{itemize}
Important: note that the data precision is 50cm per pixel. As BD Ortho is already in raster format, the only transformation we had to apply was resampling and cropping. In opposition, BD Foret and BD Topo are vector-based datasets that need to be rasterized before being used. We have used 
the \verb#gdal_rasterize# command from GDAL tools to do so. 

Initially, a large number of classes were present in the dataset. In BD Foret, a lot of information cannot be inferred from the satellite image (for example, difference between species). We reduced the number of classes by merging them and finally retained the following ones:
\begin{itemize}
    \item Sparse forest
    \item Dense forest
    \item Moor
    \item Herbaceous formation
    \item Building
    \item Road
\end{itemize}

The purpose of the two last classes is twofold. We first wanted to avoid trapping the training into false segmentation, because buildings and roads were visually present in the satellite images and were initially assigned a vegetation class. Second, the segmentation is more precise and gives more identification of the different image elements.

\subsection{Dataset preparation}
Our training and test datasets are composed of tiles prepared from IGN open data. Each tile has a 1000x1000 resolution representing a 500m x 500m footprint (the resolution is 50cm per pixel). We mainly used data from the Hautes-Alpes department, and we took spatially spaced data to have as much diversity as possible and to limit the area without information (unfortunately, some places lack information). A total of 600 tiles have been used to train the model.

The file structure of the dataset is as follows:

\begin{verbatim}
|-- data
|   |-- ign
|   |   |-- annotations
|   |   |   |-- training
|   |   |   |   |-- xxx.png
|   |   |   |   |-- yyy.png
|   |   |   |   |-- zzz.png
|   |   |   |-- validation
|   |   |-- images
|   |   |   |-- training
|   |   |   |   |-- xxx.png
|   |   |   |   |-- yyy.png
|   |   |   |   |-- zzz.png
|   |   |   |-- validation
\end{verbatim}

\begin{figure}
    \centering
    \includegraphics[width=\linewidth]{Images/caption.png}\\
    \begin{tabular}{cc}
 Original segmentation            &  Our segmentation \\
\includegraphics[width=0.45\linewidth]{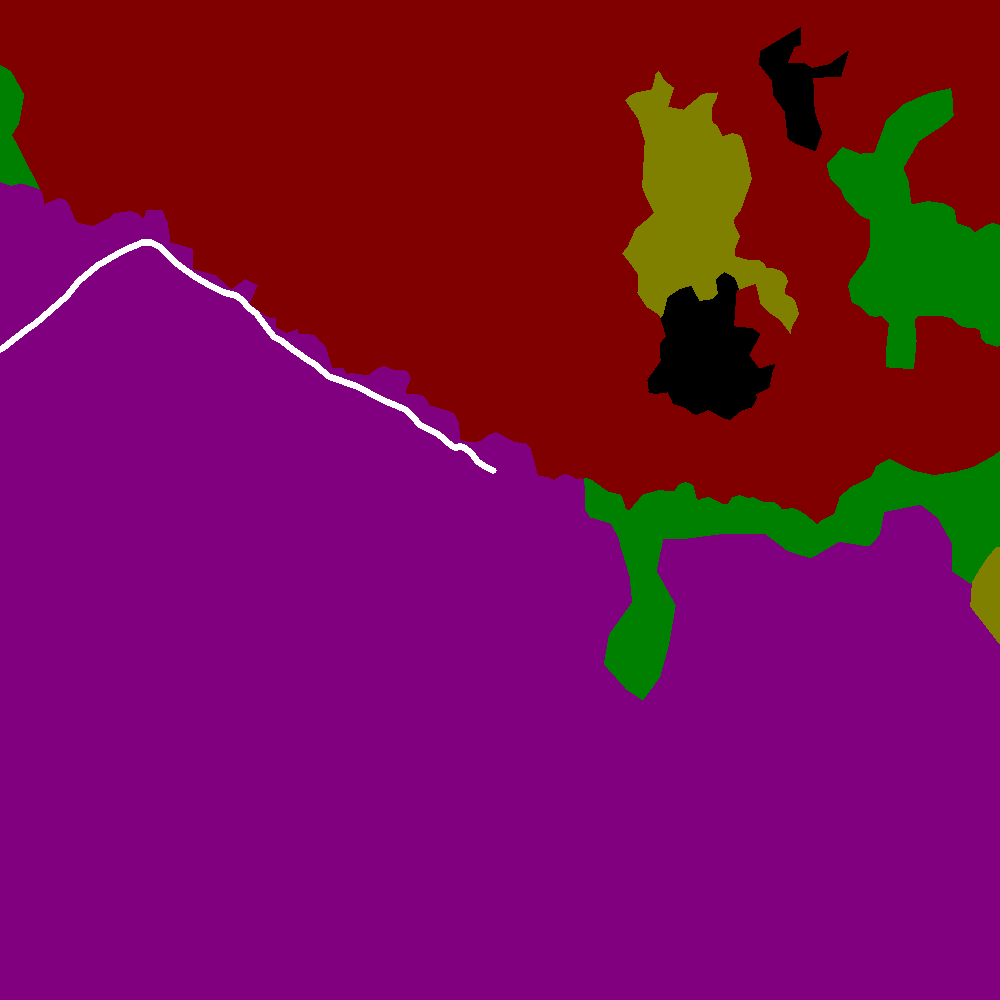}  & \includegraphics[width=0.45\linewidth]{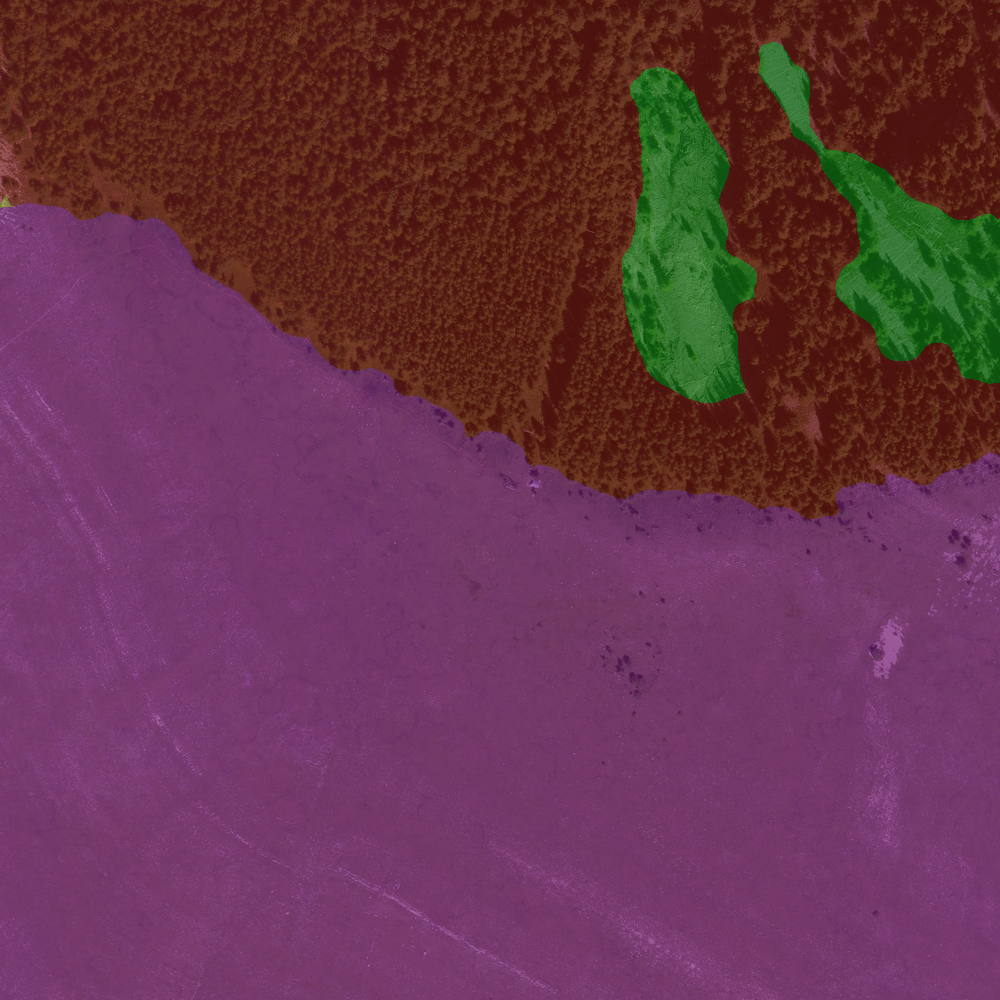}\\
\includegraphics[width=0.45\linewidth]{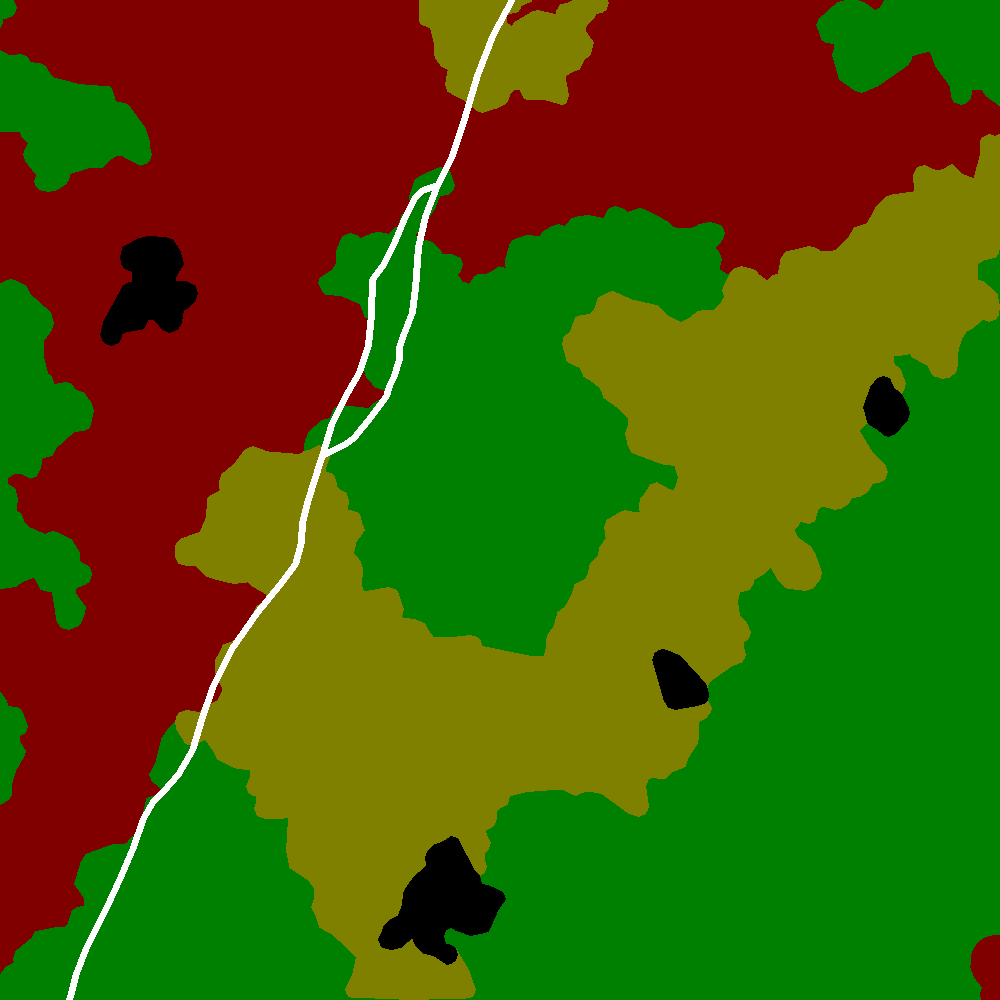} & \includegraphics[width=0.45\linewidth]{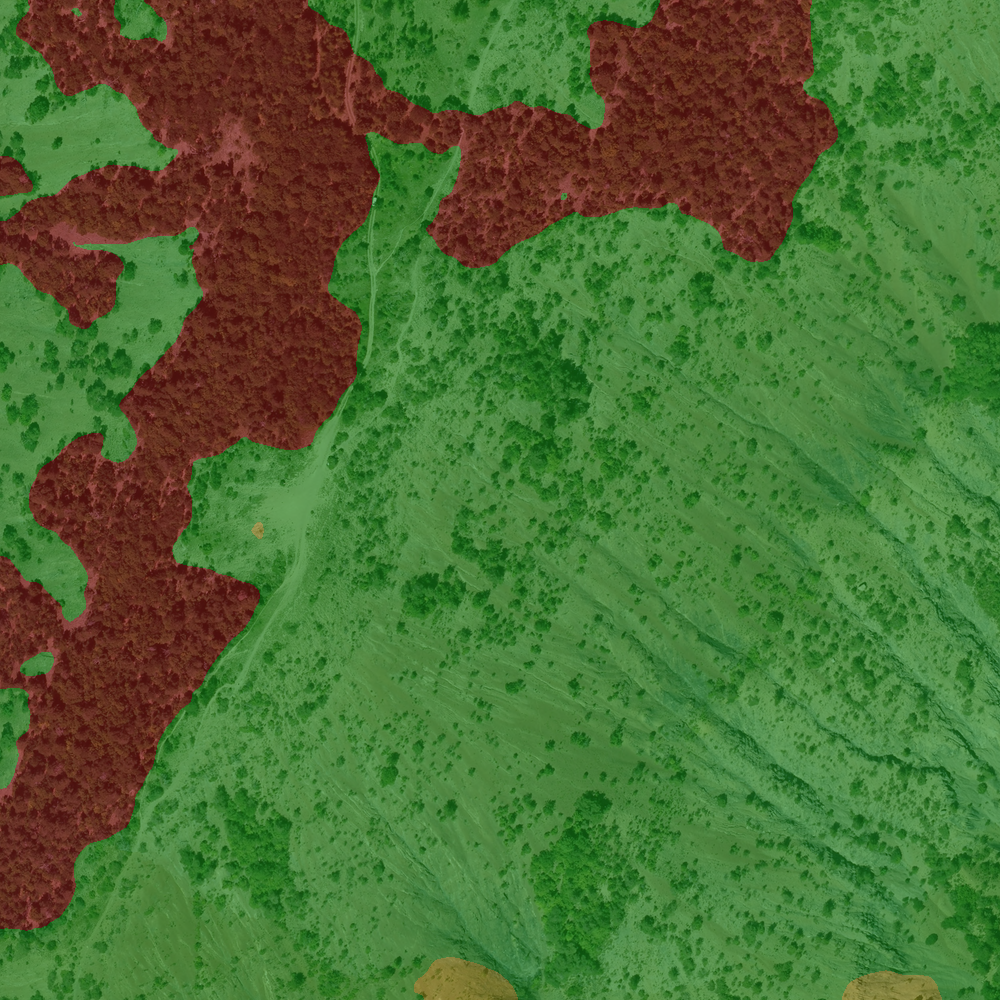}\\
\includegraphics[width=0.45\linewidth]{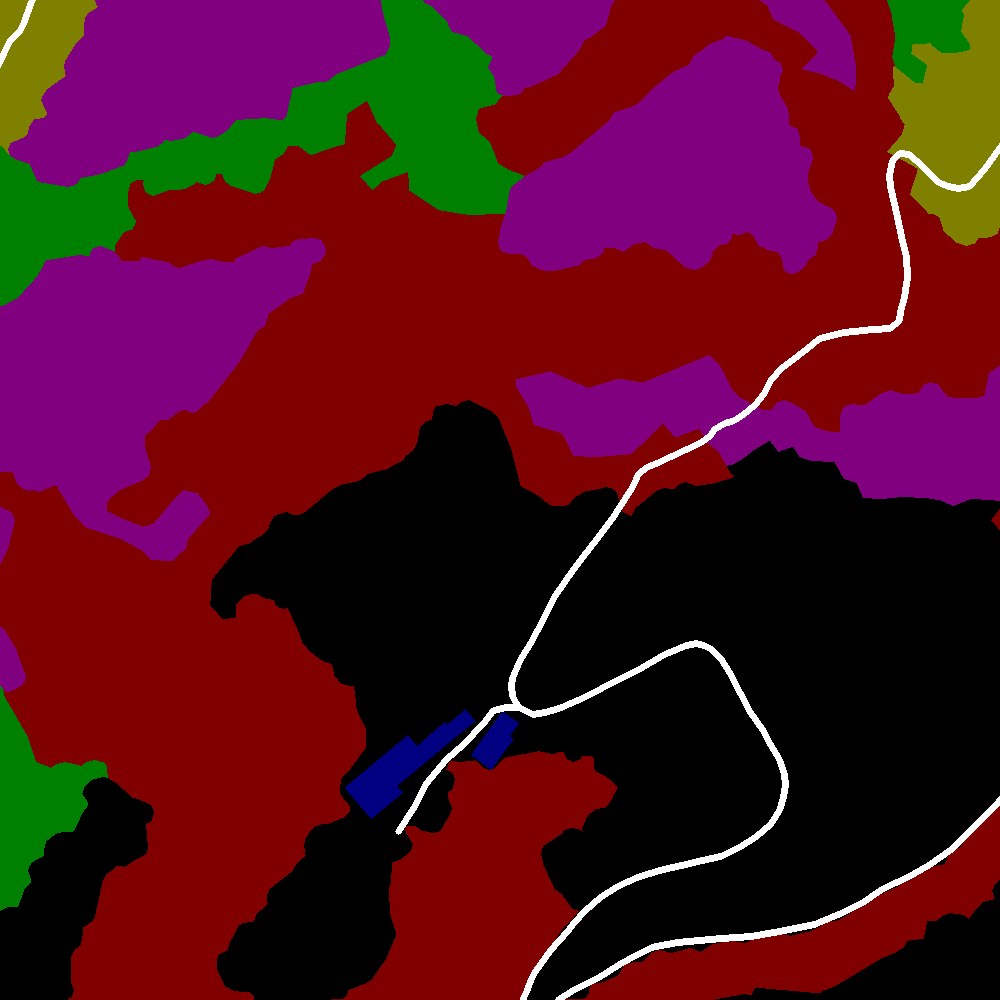} & \includegraphics[width=0.45\linewidth]{Images/c19_0935_6390.png}\\
\end{tabular}
    \caption{Main results}
    \label{fig:results}
\end{figure}

\begin{figure}
    \centering
    \includegraphics[width=\linewidth]{Images/caption.png}\\
    \begin{tabular}{cc}
    Original segmentation            &  Our segmentation \\ 
\includegraphics[width=0.45\linewidth]{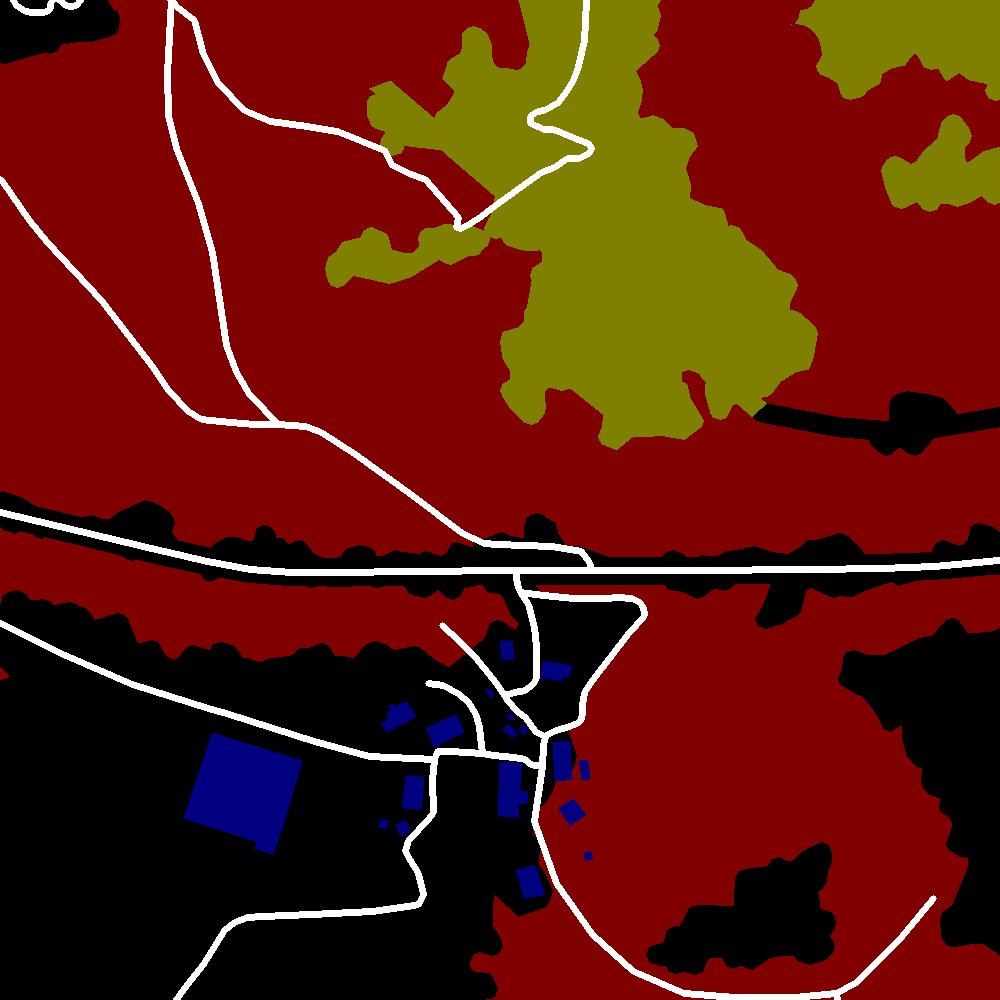}   |  & \includegraphics[width=0.45\linewidth]{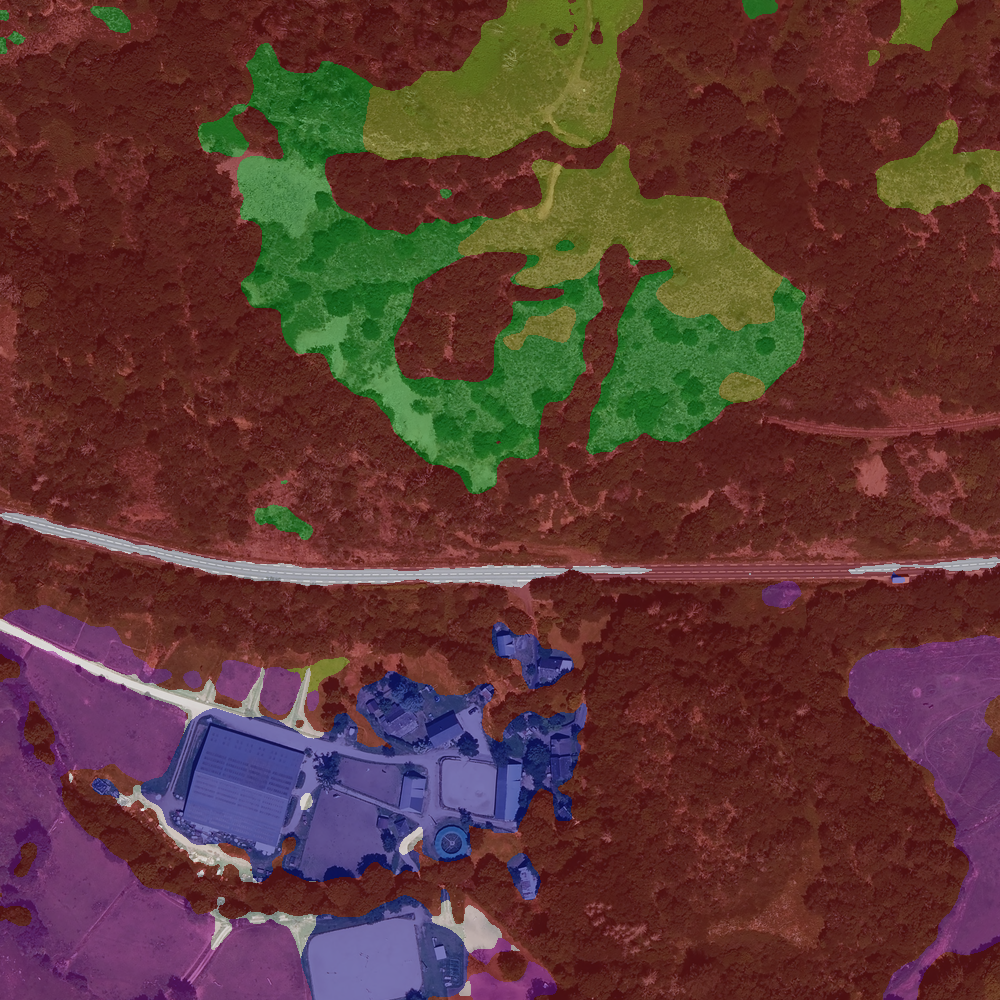} \\
\includegraphics[width=0.45\linewidth]{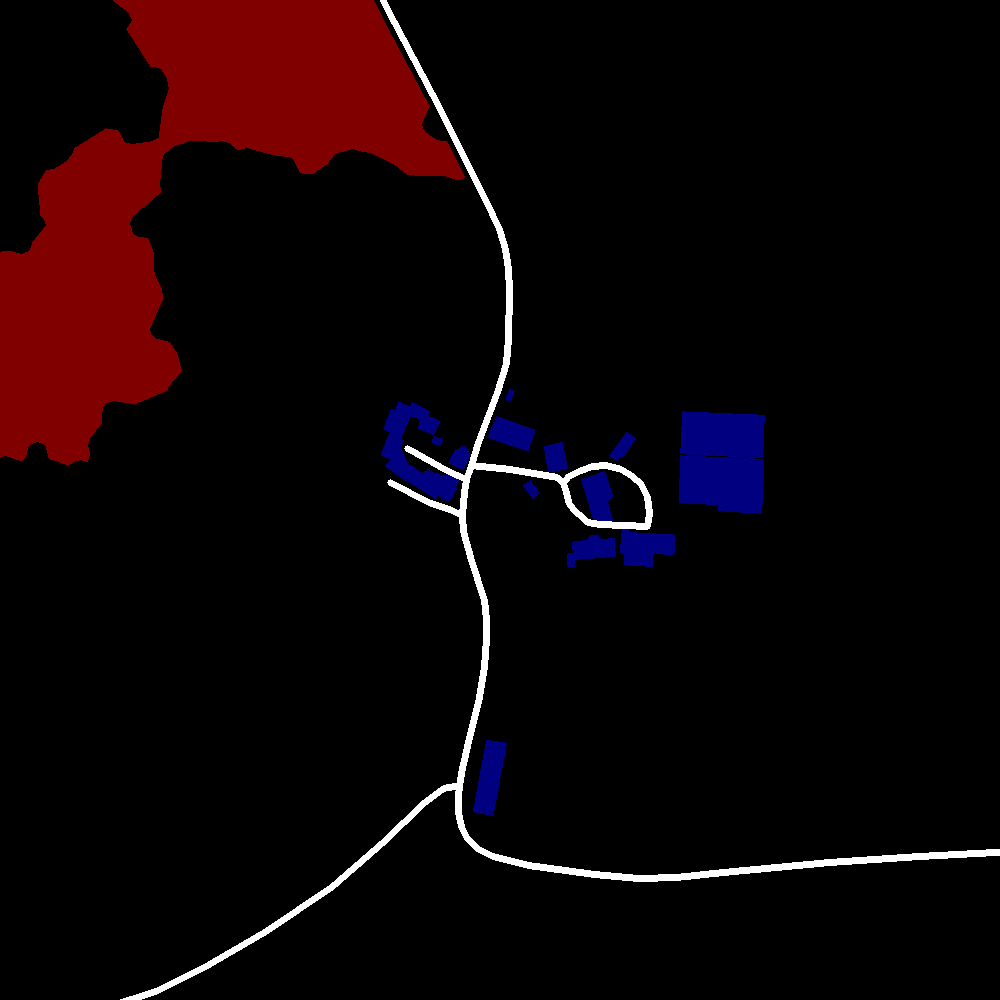}  |  & \includegraphics[width=0.45\linewidth]{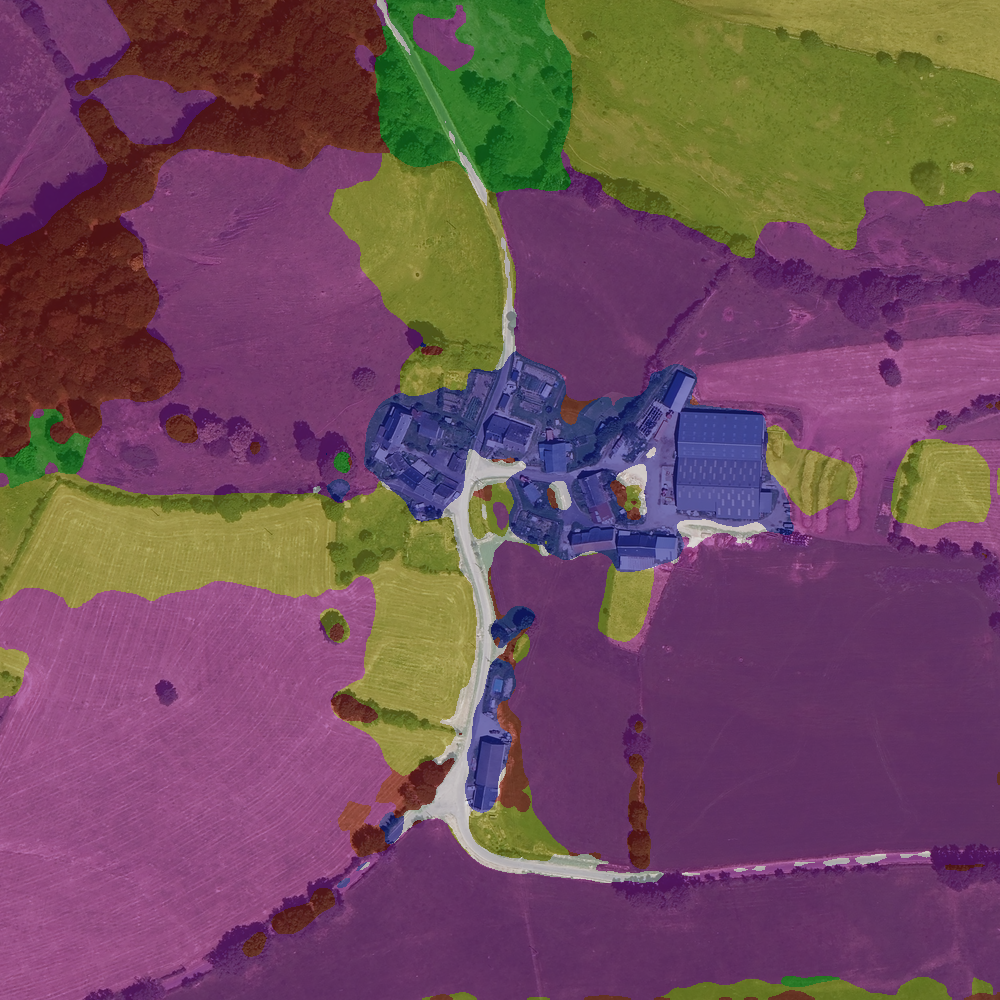} \\
\includegraphics[width=0.45\linewidth]{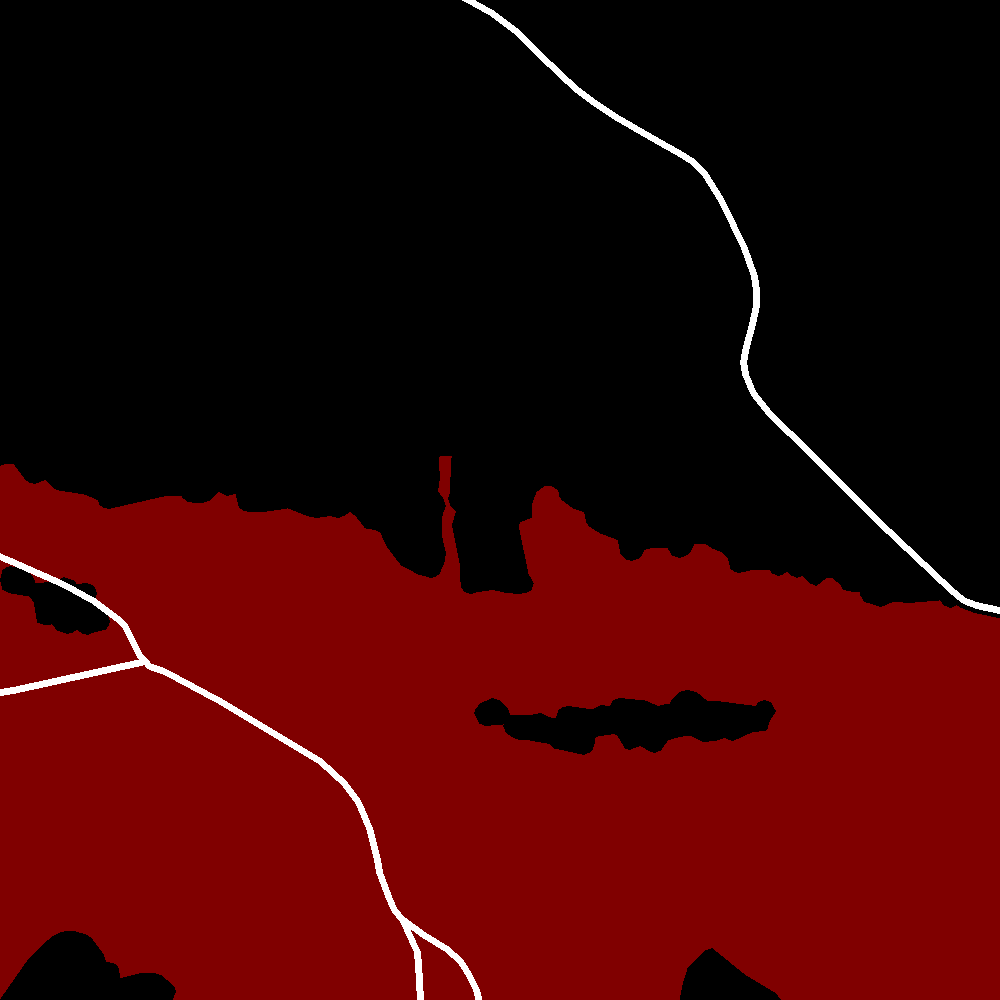}  |  & \includegraphics[width=0.45\linewidth]{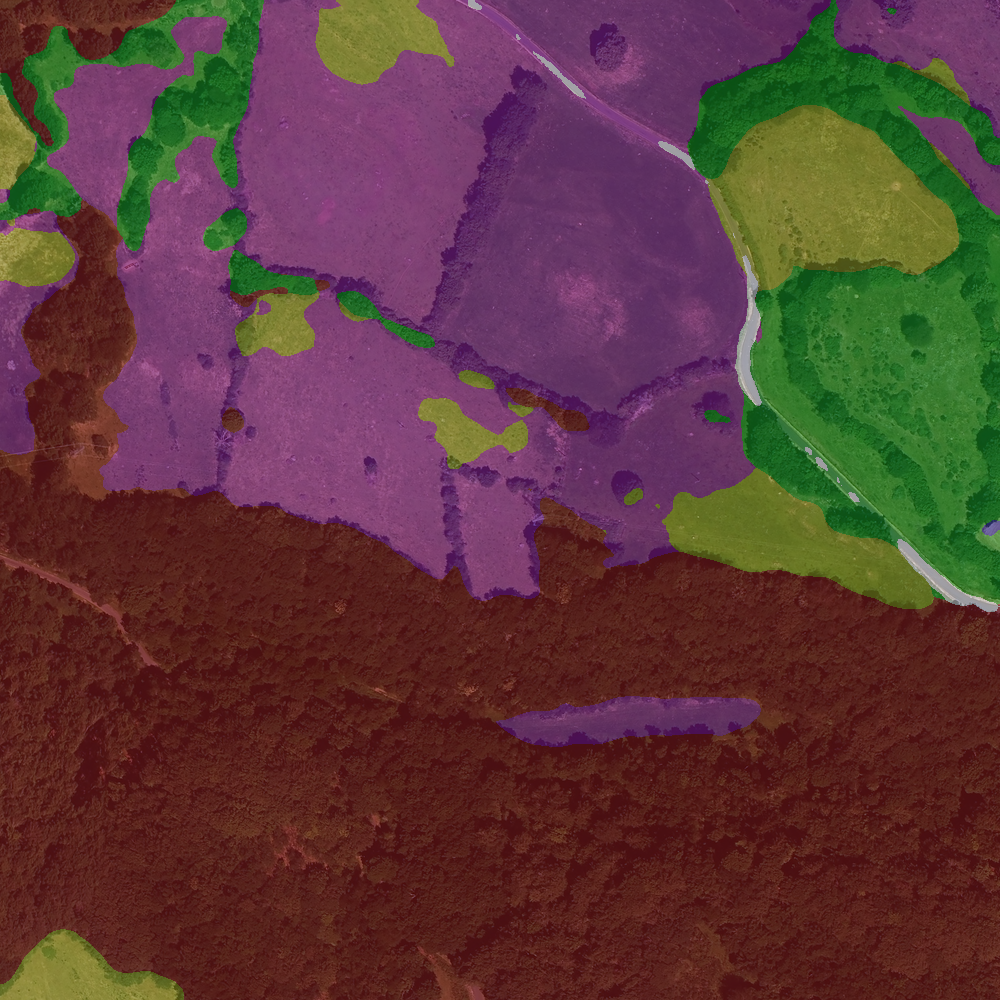} \\
\end{tabular}
    \caption{Cantal results}
    \label{fig:cantalresults}
\end{figure}

\subsection{Information on the training}
During the training, an ImageNet-22K pretrained model was used and we added weights on each class because the dataset was not balanced in classes distribution. The empirically chosen weights we have used are:
\begin{itemize}
\item Dense forest: 0.5
\item Sparse forest: 1.31237
\item Moor: 1.38874
\item Herbaceous formation: 1.39761
\item Building: 1.5
\item Road: 1.47807
\end{itemize}

\section{Experimental results}
\label{sec:results}
\begin{centering}
\begin{tabular}{|c|c|c|c|c|c|c|}\hline
     Backbone & Method & Crop Size & Lr Schd & mIoU & config & model \\\hline
     Swin-L & UPerNet & $384\times 384$ & $60$K & $54.22$ & config \footnote{\url{configs/swin/config_upernet_swin_large_patch4_window12_384x384_60k_ign.py} on the repository} & model\footnote{\url{https://drive.google.com/file/d/1EarMOBHx6meawa6izNXJUfXRCTzhKT2M/view}} \\\hline
\end{tabular}
\end{centering}

\vspace{1em}

Figure \ref{fig:results} shows some comparison between the original segmentation and the segmentation that has been obtained after the training (Hautes-Alpes dataset).

We have also tested the model on satellite photos from another French department to see if the trained model generalizes to other locations. 
We chose Cantal and a few samples of the obtained results can be seen in figure \ref{fig:cantalresults}.
These latest results show that the model is capable of producing a segmentation even if the photos are located in another department and even if there are a lot of pixels without information (in black), which is encouraging.

\subsection{Limitations}
\begin{figure}
    \centering
    \includegraphics[width=\linewidth]{Images/caption.png}\\
    \begin{tabular}{cc}
    Original segmentation            &  Our segmentation \\ 
\includegraphics[width=0.45\linewidth]{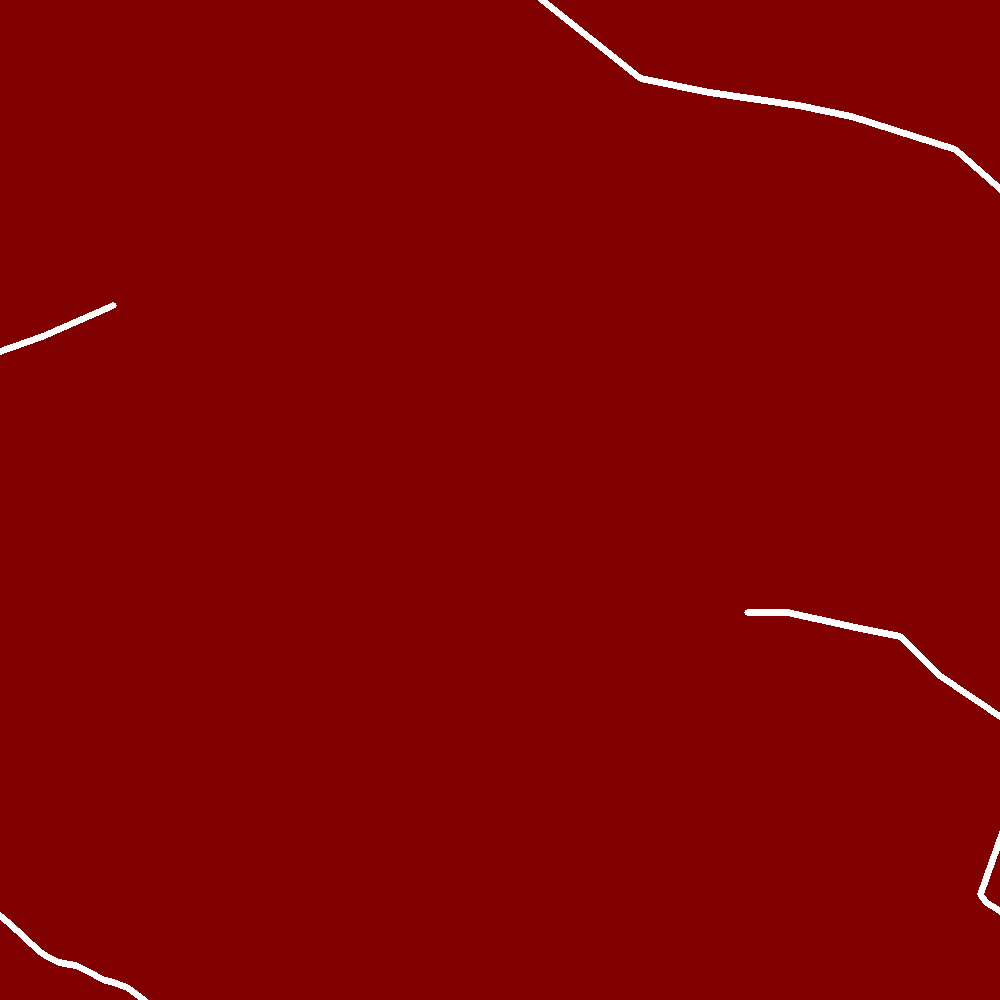}   |  & \includegraphics[width=0.45\linewidth]{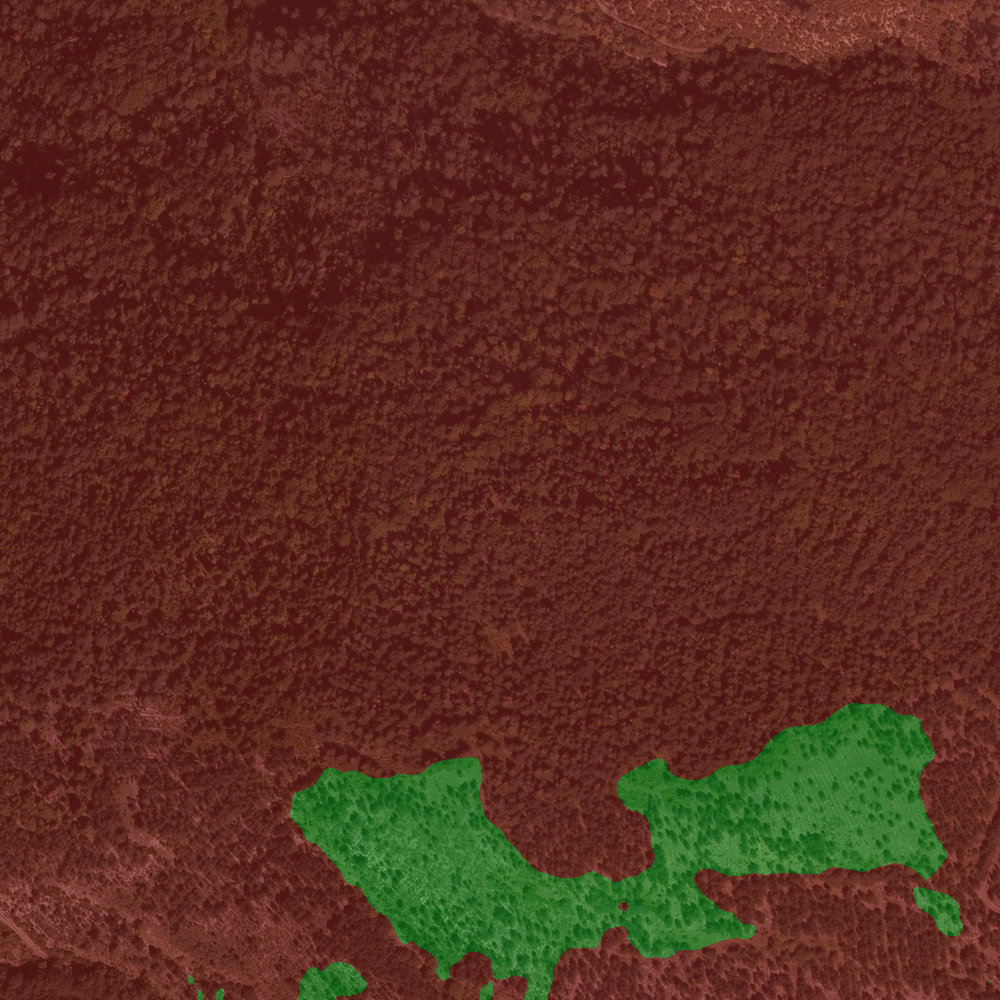} \\
\end{tabular}
    \caption{Example of limitation}
    \label{fig:limitation}
\end{figure}

As illustrated in the previous images, the results are not perfect. This is caused by the inherent limits of the data used during the training phase. The main limitations are:  
\begin{enumerate}
    \item The satellite photos and the original segmentation were not made at the same time, so the segmentation is not always accurate. For example, we can see in figure \ref{fig:limitation} a zone is segmented as "dense forest" even if there are not many trees (that is why the segmentation after training, on the right, classed it as "sparse forest").
    \item Sometimes there are zones without information (represented in black) in the dataset. Fortunately, we can ignore them during the training phase, but we also lose some information, which is a problem: we thus removed the tiles that had more than 50\% of unidentified pixels to try to improve the training.
    \item Road segmentation is not accurate because sometimes the information is not visible in the image (hidden by trees for example), which obviously prevents it from being detected.
\end{enumerate} 

\section{Repository}
The source code, which is only a fork from the implementation of Swin Transformer can be found on github together with usage details\footnote{\url{https://github.com/koechslin/Swin-Transformer-Semantic-Segmentation}}.

\section{Acknowledgments}
This work has been funded by the ANR, project Ampli ANR-20-CE23-0001.

\bibliographystyle{plain}
\bibliography{biblio}

\end{document}